\documentclass[conference,10pt,twocolumn]{IEEEtran}
\IEEEoverridecommandlockouts
\usepackage{cite}
\usepackage{amsmath,amssymb,amsfonts}
\usepackage{graphicx}
\usepackage{textcomp}
\usepackage{xcolor}
\usepackage{booktabs}
\usepackage{float}
\usepackage{algorithm,algorithmicx,algpseudocode}
\usepackage{multirow}
\usepackage{enumitem}
\usepackage{mathtools}
\usepackage[letterpaper, left=0.75in, right=0.75in, top=1in, bottom=1in]{geometry}
\usepackage{caption}
\usepackage{subcaption}
\usepackage{titlesec} 
\usepackage{siunitx} 

\DeclareCaptionLabelFormat{bold}{\textbf{#1 #2}}
\renewcommand{\thetable}{\arabic{table}}

\renewcommand{\thesection}{\arabic{section}}
\renewcommand{\thesubsection}{\thesection.\arabic{subsection}}

\titleformat{\section}{\normalfont\bfseries\normalsize}{\thesection}{1em}{}
\titleformat{\subsection}{\normalfont\bfseries\normalsize}{\thesubsection}{1em}{}

\renewcommand{\baselinestretch}{1}

\def\BibTeX{{\rm B\kern-.05em{\sc i\kern-.025em b}\kern-.08em
    T\kern-.1667em\lower.7ex\hbox{E}\kern-.125emX}}

\usepackage{setspace}  

\newenvironment{customabstract}
  {\par\setlength{\parindent}{0pt} 
   \textbf{\uppercase{Abstract}}\par}
  {\par\vspace{2ex}} 

\begin{document}

\title{\fontsize{18}{22}\selectfont Democratizing MLLMs in Healthcare: TinyLLaVA-Med for Efficient Healthcare Diagnostics in Resource-Constrained Settings}

\author{
    \IEEEauthorblockN{
        Aya El Mir\IEEEauthorrefmark{1}, 
        Lukelo Thadei Luoga\IEEEauthorrefmark{1}, 
        Boyuan Chen, 
        Muhammad Abdullah Hanif, 
        Muhammad Shafique
    }
    \IEEEauthorblockA{\textit{eBrain Lab, Division of Engineering, New York University Abu Dhabi, UAE\IEEEauthorrefmark{2}} \\
    \{ae2195, ltl2113, bc3194, mh6117, muhammad.shafique\}@nyu.edu \vspace{-10mm}}
     \thanks{\IEEEauthorrefmark{1}Equal contribution}
}

\maketitle

\begin{customabstract}
Deploying Multi-Modal Large Language Models (MLLMs) in healthcare is hindered by their high computational demands and significant memory requirements, which are particularly challenging for resource-constrained devices like the Nvidia Jetson Xavier. This problem is particularly evident in remote medical settings where advanced diagnostics are needed but resources are limited. In this paper, we introduce an optimization method for the general-purpose MLLM, TinyLLaVA, which we have adapted and renamed TinyLLaVA-Med. This adaptation involves instruction-tuning and fine-tuning TinyLLaVA on a medical dataset by drawing inspiration from the LLaVA-Med training pipeline. Our approach successfully minimizes computational complexity and power consumption, with TinyLLaVA-Med operating at 18.9W and using 11.9GB of memory, while achieving accuracies of 64.54\% on VQA-RAD and 70.70\% on SLAKE for closed-ended questions. Therefore, TinyLLaVA-Med achieves deployment viability in hardware-constrained environments with low computational resources, maintaining essential functionalities and delivering accuracies close to state-of-the-art models.
\end{customabstract}


\section{INTRODUCTION}
\label{Sec1:Introduction}
The transformative potential of AI in healthcare is increasingly recognized, primarily for enhancing diagnostic accuracy and personalizing care\cite{bekbolatova2024transformative}\cite{poalelungi2023advancing} \cite{jiang2017artificial}\cite{shaheen2021applications}. In healthcare, a domain characterized by diverse data forms such as medical images, textual reports, and real-time sensor data, AI technologies that can effectively handle and utilize this multimodal information are crucial \cite{acosta2022multimodal}. These technologies not only improve clinical decision-making but also enable comprehensive patient management, thus optimizing health outcomes. Moreover, AI applications extend from reducing routine administrative burdens to supporting complex diagnostic processes, thereby increasing healthcare delivery efficiency and patient-centered care. \cite{mesko2023impact}

In response to the critical need for AI technologies that can handle multimodal data in healthcare, several multimodal large language models (MLLMs) like LLaVA-Med\cite{li2024llava}, Med-PaLM \cite{singhal2023towards}, Med-flamingo,\cite{moor2023med}, PubMedCLIP\cite{eslami2023pubmedclip} and  BiomedCLIP\cite{zhang2023biomedclip} have been proposed. These MLLMs integrate Large Language Models (LLMs) with Vision Encoders, thus possessing capabilities that extend beyond textual understanding and analysis to include image processing capabilities. This enables them to simultaneously interpret both textual data and medical images, facilitating more accurate and comprehensive diagnostics and decision-making in healthcare. By rapidly processing and synthesizing diverse data types, these models can significantly advance patient care, enabling quicker, more precise diagnoses and personalized treatment plans, thus, transforming healthcare into a more efficient, effective, and patient-centered service \cite{acosta2022multimodal}\cite{mesko2023impact}.

However, the deployment of these models in practical settings is constrained by their large size and substantial computational requirements. This becomes a significant barrier in resource-limited environments, typical in remote or underserved areas, limiting access to state-of-the-art AI medical technologies. These regions often lack high-performance computing (HPC) facilities and powerful GPUs needed to run large multimodal models, which typically require substantial memory and computational resources. By proposing a model with significantly fewer parameters, we make it feasible to run on less powerful hardware, such as embedded devices like the Nvidia Jetson Xavier. This reduction in resource requirements makes advanced AI diagnostics more accessible in these regions, bridging the gap between technological capability and accessibility where it is most needed.

Our work proposes TinyLLaVA-Med, a compact multimodal large language model (MLLM) developed by fine-tuning the general-purpose TinyLLaVA on medical datasets using the training framework of LLaVA-Med MLLM. TinyLLaVA-Med MLLM is designed to be deployable on embedded systems with low computational power, such as the Nvidia Jetson Xavier. While existing studies have shown that smaller multimodal large language models (MLLMs) like MoE-TinyMed\cite{jiang2024moe} can achieve or even surpass the accuracy of larger models in medical settings, they did not focus on the practical deployment of these models on resource-constrained devices. Our research not only confirms that TinyLLaVA-Med attains high accuracy but also extends these findings by demonstrating the practical deployment of this model on embedded devices. This step showcases the potential of implementing advanced AI-driven medical diagnostics in environments where computational resources are significantly limited, emphasizing the potential of MLLMs to revolutionize healthcare delivery in remote and underserved areas.

\section{BACKGROUND AND RELATED WORK}
Understanding the current state of the art in Multimodal Large Language Models (MLLMs) used in healthcare is crucial for identifying gaps and opportunities for innovation. This section highlights significant developments in MLLMs that enhance medical diagnostics and patient care, leveraging diverse data forms such as text and images. We also detail the architecture and training methodology of TinyLLaVA, a model designed to bring these advanced capabilities to resource-constrained environments, illustrating our project's basis and its alignment with leading practices in the field.

\subsection{Multimodal Large Language Models in Healthcare} 
Recent advancements in multimodal large language models (MLLMs) have significantly enhanced medical diagnostics and patient care. These models integrate diverse data types, including text and images, to boost both the accuracy and efficiency of diagnostics. MLLMs frequently contain three key components: a pre-trained modality encoder, a pre-trained large language model (LLM), and a modality interface. The modality encoder, often a convolutional neural network (CNN) or transformer-based model, processes visual data such as X-ray, MRI, and CT scan images, extracting and enhancing detailed features. The LLM handles text generation, interpreting and producing medical reports, patient records, and research articles, drawing from extensive medical literature to provide precise, context-rich interpretations. The modality interface ensures seamless integration of these components, using techniques like cross-modal attention mechanisms to align text and image data effectively \cite{acosta2022multimodal}\cite{li2024llava}\cite{eslami2023pubmedclip}\cite{singhal2023towards}. This structure allows MLLMs to produce comprehensive and accurate diagnostic conclusions, improving patient outcomes by integrating detailed image analysis with contextual textual information.
Leading examples of Multimodal Large Language Models (MLLMs) in healthcare include LLaVA-Med, which leverages a large-scale biomedical dataset for conversational support on biomedical images \cite{li2024llava}, and Med-PaLM 2, renowned for its physician-level accuracy in medical question answering due to extensive domain-specific fine-tuning \cite{singhal2023towards}. Med-Flamingo introduces adaptability through few-shot learning, proficiently managing real-time medical visual question answering \cite{moor2023med}. Additionally, BiomedCLIP and PubMedCLIP excel in biomedical image-text pair analysis, significantly enhancing diagnostic precision with their specialized training datasets \cite{zhang2023biomedclip}\cite{eslami2023pubmedclip}. These MLLMs can be used in real-world scenarios to provide clinical decision support in medical fields such as radiology and pathology, thereby showcasing their potential to improve diagnostic accuracy and patient outcomes\cite{acosta2022multimodal}.

\subsection{TinyLLaVA} 
\begin{figure}[ht]
  \centering
  \includegraphics[width=0.4\textwidth, trim={1cm 2cm 1cm 3cm}, clip]{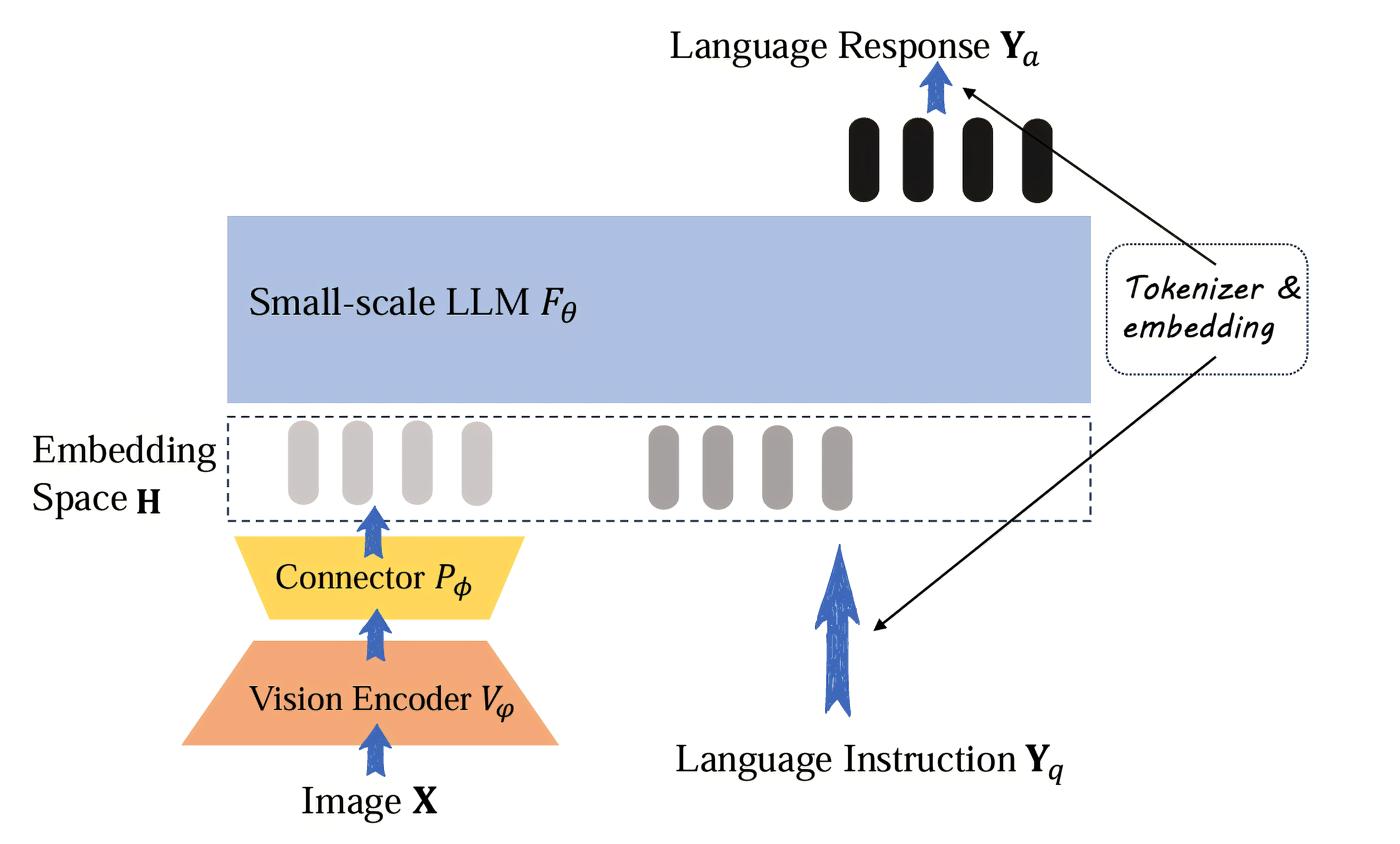}
  \caption{TinyLLaVA Architecture \cite{zhou2024tinyllava}}
  \label{fig:tinyllava}
\end{figure}

The TinyLLaVA model represents a significant advancement in the field of Multimodal Small Language Models (MSLMs), developed to offer a cost-effective and computationally efficient alternative to larger models without compromising on performance. Illustrated in Figure \ref{fig:tinyllava}, the architecture integrates three primary components: a vision encoder \( V_\phi \), a small-scale language model \( F_\theta \), and a connector \( P_\phi \). The vision encoder processes images into visual patch features, while the language model handles textual data to generate responses. The connector aligns these visual and textual elements within the embedding space, facilitating coherent multimodal interaction. TinyLLaVA was trained using a unique approach that involves two primary stages: pre-training and supervised fine-tuning. During pre-training, the model was trained to align the vision and text information in the embedding space using image-caption style data formats. This stage was crucial for preparing the model's layers to handle real-world data by aligning different modalities effectively. The supervised fine-tuning stage then utilized image-text pair data in a multi-turn conversation format, optimizing the model's responses to be contextually relevant and accurate. This model not only supports efficient processing but also maintains competitive performance, making it ideal for applications requiring robust multimodal understanding in resource-limited settings \cite{zhou2024tinyllava}.

\section{METHODOLOGY}
\begin{figure}[ht]
  \centering
  \includegraphics[width=\linewidth]{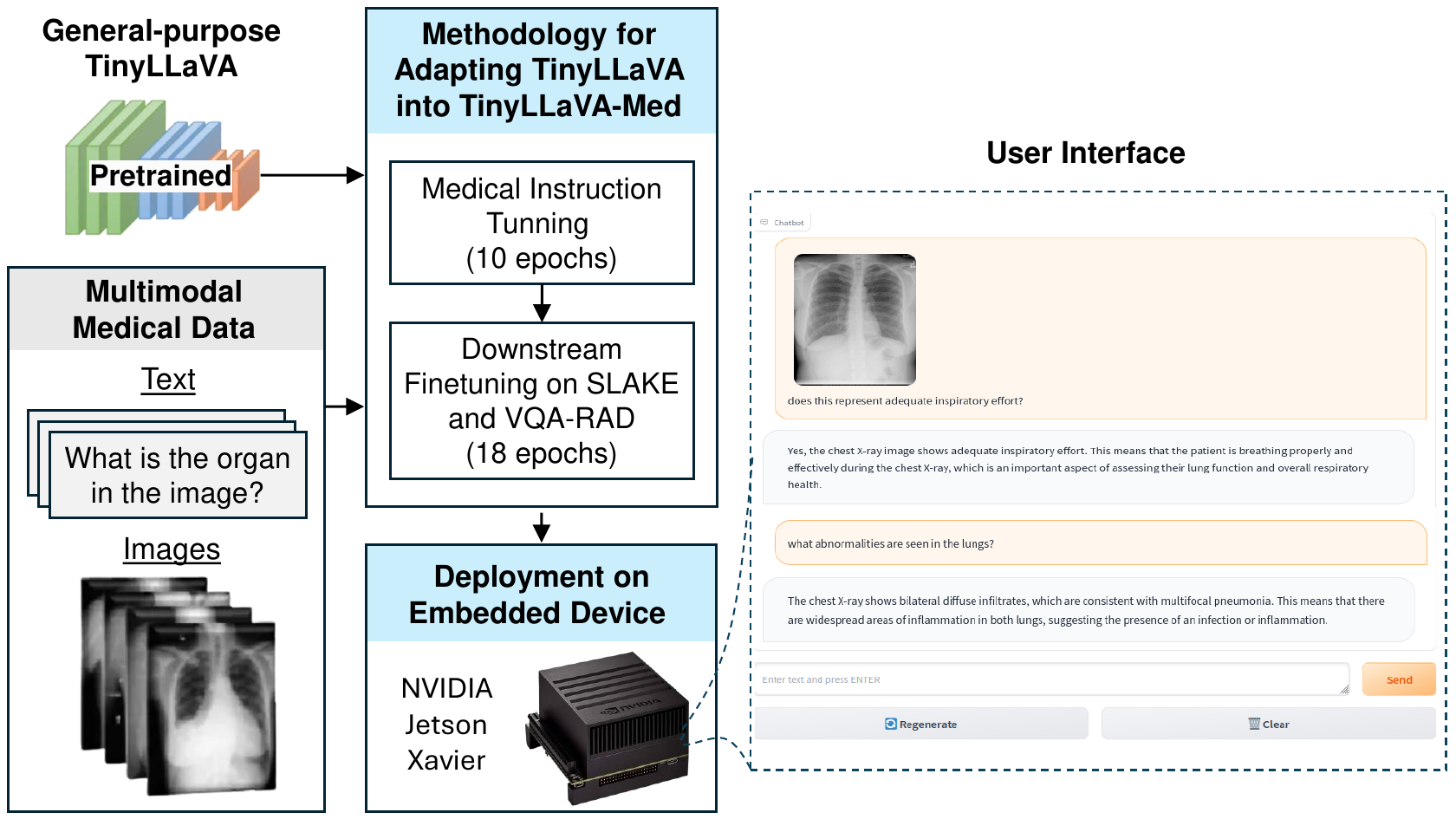} 
  \caption{Flowchart illustrating the methodology of adapting TinyLLaVA into TinyLLaVA-Med for deployment on embedded devices.}
  \label{fig:methodology}
  \vspace{-0.05em} 
\end{figure}

The methodology to adapt the TinyLLaVA model for medical applications involved a sequential approach, beginning with instruction-tuning and downstream fine-tuning and ending with deployment on an embedded device. Figure \ref{fig:methodology} outlines the entire process.


\subsection{Instruction-Tuning}
Beginning with the pretrained general-purpose TinyLLaVA model, our first step was to adapt it to interpret and process multimodal medical data that integrates text and imagery. This adaptation drew inspiration from the LLaVA-Med model's approach\cite{li2024llava}, which involved tuning a pretrained model specifically for medical applications. The Instruction Tuning stage mirrored the LLaVA-Med’s second stage. In this phase, the TinyLLaVA model underwent end-to-end tuning to enhance its ability to follow diverse instructions and perform tasks within a conversational medical context. The tuning focused on updating the projection layer and language model weights, while the visual encoder weights remained unchanged. For this tuning, we employed the Biomedical Instruction-Tuning Data from LLaVA-Med, sourced from PMC-15M\cite{zhang2023large}, containing 60,000 image-text pairs across major imaging modalities. Dataset preparation involved enhancing captions with sentences from PubMed articles and using GPT-4\cite{achiam2023gpt} to generate multi-round conversational data, refining the model's capability for detailed medical dialogues. The Instruction tunning of TinyLLaVA significantly improved the model’s capacity to interact within medical contexts, transforming TinyLLaVA into TinyLLaVA-Med.

\subsection{Fine-tuning to Downstream Datasets} 
Following instruction tuning, TinyLLaVA-Med underwent downstream fine-tuning on specialized biomedical Visual Question Answering (VQA) datasets, such as VQA-RAD\cite{lau2018dataset} and SLAKE\cite{liu2021slake}. These datasets are critical for evaluating the effectiveness of our training pipeline, as they contain both open-ended and close-ended medical questions, serving as a benchmark for assessing the model's performance. This fine-tuning step helped to attain a highly accurate and dataset-specific TinyLLaVA-model.

\subsection{Deployment on Embedded Device}
The final step involved deploying TinyLLaVA-Med on the Nvidia Jetson Xavier, an embedded device chosen for its balance of computational power and energy efficiency suitable for real-time applications in healthcare. This deployment tested the model’s operational effectiveness, particularly its ability to process data swiftly and accurately in an embedded system environment, thereby confirming its readiness for practical medical use.

Each phase of this methodology not only refined TinyLLaVA’s capabilities but also ensured that the final model, TinyLLaVA-Med, was robust and efficient enough to function in resource-limited healthcare environments.

\section{RESULTS}

\subsection{Datasets}
\color{black}

In the instruction-tuning stage, our TinyLLaVA-Med leverages the LLaVA-Med\cite{li2024llava} open-sourced dataset, containing 60K image-text pairs from five major imaging domains, including Chest X-ray, MRI, Histology, Gross pathology, and CT. This dataset provided a core foundation for our work. Our initial task involved downloading and extracting images from PMC-15M articles to ensure all necessary images were included in our dataset. This process ensured that our model would effectively learn from the well-structured and detailed multimodal biomedical dataset.

\subsection{Evaluation Metrics for TinyLLaVA-Med}
To assess whether TinyLLaVA-Med can reach the required medical accuracy and be successfully deployable on the Nvidia Jetson Xavier board, we employ the following metrics. These are designed to measure both the model's effectiveness in medical applications and its operational efficiency within the hardware constraints of the Jetson Xavier, ensuring that it performs optimally in real-world healthcare environments.

\subsubsection{Medical Capability Metrics}
We evaluate the diagnostic capabilities of TinyLLaVA-Med using the VQA-RAD and SLAKE datasets, both specialized biomedical Visual Questioning (VQA) benchmarks in healthcare. The VQA-RAD dataset includes 315 radiology images and 3515 QA pairs covering various body parts and question types like abnormality and modality\cite{lau2018dataset}, while the SLAKE dataset contains 642 images and over 7000 QA pairs annotated by physicians, featuring semantic segmentation masks and enhancements from an external medical knowledge graph\cite{liu2021slake}. These datasets allow for a comprehensive assessment of TinyLLaVA-Med’s performance in medical image understanding and question answering, using metrics such as recall for open-ended questions and accuracy for closed-ended questions to gauge the model's ability across different medical scenarios.

\subsubsection{Hardware Deployment Evaluation Metrics}
\paragraph{GPU Utilization Efficiency}
Our goal is to optimize GPU utilization on the Nvidia Jetson Xavier to nearly 100\% capacity. High utilization indicates that the model maximizes the available computational resources, which is crucial for efficient operation and rapid response times needed in real-time healthcare applications. Low utilization may suggest that the model either requires minimal computation or is not fully optimized for the hardware.

\paragraph{Power Efficiency}
Power efficiency is assessed by ensuring the model operates within Jetson Xavier's power consumption range of 10W-30W. This criterion supports deployment in medical settings with limited power availability, ensuring the model remains functional without excessive energy use.

\paragraph{Memory Footprint}
Optimizing the memory footprint is crucial for the Jetson Xavier, which has 32GB of 256-bit LPDDR4x memory. Our goal is to ensure TinyLLaVA-Med operates within this memory limit to support high throughput for critical tasks like real-time medical imaging and diagnostics without exceeding system capacity.

\subsection {Results}
\subsubsection{Instruction-Tuning Training Stage Results}
\begin{figure}[ht]
  \centering
  \includegraphics[width=\linewidth]{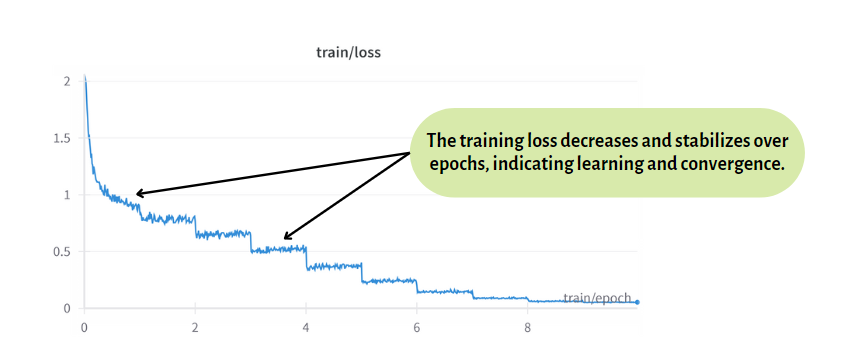} 
  \caption{Training loss of TinyLLaVA-Med on PMC-15M dataset over epochs, indicating effective learning and model convergence during the instruction tuning stage.}
  \label{fig:tinyllava-med-training-loss}
\end{figure}

The instruction tuning stage involved end-to-end fine-tuning where the model was adjusted to follow specific instructions and perform tasks within a conversational medical context. During the stage, we monitored the training loss as seen in Figure \ref{fig:tinyllava-med-training-loss}.

The training loss for TinyLLaVA-Med on the medical dataset demonstrates significant learning and adaptation as the loss sharply declined from initial values, indicating rapid adaptation to the training data. This was followed by a stabilization phase, where the loss plateaued, reflecting the model's convergence towards minimal loss. These results indicate effective learning and model stability.  

\subsubsection{Hardware Performance Results}
To assess the operational efficiency of TinyLLaVA-Med, we conducted hardware performance monitoring of the Nvidia Jetson AGX Xavier during the inference of TinyLLaVA-Med. The following metrics were noted and are summarized in Table \ref{tab:hardware-performance-metrics-tinyllava-med}:
\begin{itemize}
  \item \textbf{GPU Utilization}: Our model achieved a GPU utilization rate of 62\%. Typically, higher utilization closer to 100\% is preferable for maximizing computational resources, especially in performance-critical applications like real-time healthcare diagnostics. However, the 62\% rate suggests that while the system is not overloaded, there may be room to further optimize the model to use the available GPU resources more effectively.
  \item \textbf{Power Efficiency}: The power consumption was measured at 18.9W, which falls within the operational power range of 10W-30W set for the Nvidia Jetson Xavier. This confirms the model's efficient power usage.
  \item \textbf{Memory Footprint}: Memory utilization was optimized to 11.9GB out of 30.3GB RAM and 1.1GB out of 4.2GB GPU memory. This optimization reflects a significant reduction in memory usage while maintaining robust model performance.
\end{itemize}

\begin{table}[ht]
  \centering
  \setlength{\tabcolsep}{10pt} 
  \renewcommand{\arraystretch}{1.2} 
  \scriptsize 
  \caption{Performance metrics of TinyLLaVA-Med during inference on NVIDIA Jetson AGX Xavier compared to the expected values based on the Jetson operational limits.}
  \label{tab:hardware-performance-metrics-tinyllava-med}
  \begin{tabular}{|c|c|c|}
    \hline
    \multicolumn{1}{|c|}{\textbf{Metric}} & \multicolumn{1}{|c|}{\textbf{TinyLLaVA-Med}} & \multicolumn{1}{|c|}{\textbf{Expected}} \\ \hline
    GPU Utilization (\%) & 62 & Close to 100 \\ \hline
    Power Consumption (W) & 18.9 & 10-30 \\ \hline
    Memory Usage (GB) & 11.9 & 32 \\ \hline
  \end{tabular}
\end{table}

These results validate our model's design and optimization processes, confirming that TinyLLaVA-Med not only meets but exceeds the necessary operational standards for effective deployment in medical settings. This ensures efficient data processing without compromising the performance required for real-time medical diagnostics.

\subsubsection{Comparison with State-of-the-Art}
To position TinyLLaVA-Med in the current landscape of multimodal large language models (MLLMs), we compare its performance against several leading models such as LLaVA-Med (Llama7B and Vicuna7B) and TinyMoE-Med variants. These models represent the state-of-the-art in combining textual and visual data to address complex question answering tasks in medical domains, making them relevant benchmarks for our evaluation. As seen in Table \ref{tab:tinyllava_comparison}, The model displayed varied performance, excelling particularly in closed-ended questions with accuracies reaching 70.70\% in SLAKE and 64.54\% in VQA-RAD, which suggests its proficiency in scenarios requiring definitive binary answers. In comparison, open-ended question handling proved more challenging, with lower recall rates of 61.62\% and 29.85\% respectively on SLAKE and VQA-RAD.

Despite its lower accuracy in open-ended questions compared to other models, TinyLLaVA-Med's performance in closed-ended scenarios highlights its potential for specific applications where concise, binary outputs are required. The modest gap in performance between TinyLLaVA-Med and larger models such as LLaVA-Med and TinyMoE-Med illustrates the feasibility of achieving a balance between model size and accuracy. This balance is critical for deploying efficient yet capable models on platforms with limited computational resources like the Nvidia Jetson Xavier, indicating promising avenues for optimization and targeted application in real-world settings.

These results shows the potential to refine TinyLLaVA-Med further, enhancing its capability for open-ended questions while maintaining its efficiency for closed-ended tasks. Optimizing this balance can make TinyLLaVA-Med a practical solution in healthcare settings, particularly in remote or resource-constrained environments where advanced diagnostic support is crucial but computational resources are limited.

\begin{table*}[t] 
  \centering
  \setlength{\tabcolsep}{10pt} 
  \renewcommand{\arraystretch}{1.2} 
  \scriptsize 
  \caption{Performance comparison of TinyLLaVA-Med with state-of-the-art models on VQA-RAD and SLAKE datasets. This table presents a tabular comparison highlighting the method’s performance on both open-ended and closed-ended questions, contrasting it against established models.}
  \label{tab:tinyllava_comparison}
  \begin{tabular}{|c|c|c|c|c|}
    \hline
    \multirow{2}{*}{\centering\textbf{Method}} & \multicolumn{2}{|c|}{\textbf{VQA-RAD}} & \multicolumn{2}{|c|}{\textbf{SLAKE}} \\ \cline{2-5}
    & \textbf{Open (\%)} & \textbf{Closed (\%)} & \textbf{Open (\%)} & \textbf{Closed (\%)} \\ \hline
    LLaVA-Med (Llama7B) & 61.52 & 84.19 & 83.08 & 85.34 \\ \hline
    LLaVA-Med (Vicuna7B) & 64.39 & 81.98 & 84.71 & 83.17 \\ \hline
    TinyMoE-Med (Phi2-2.7Bx4:3.6B) & 52.55 & 84.56 & 85.27 & 86.78 \\ \hline
    TinyMoE-Med (StableLM-1.6Bx4:2.0B) & 47.26 & 83.82 & 82.28 & 84.86 \\ \hline
    TinyLLaVA-Med (TinyLLaVA-1.5B) & 29.85 & 64.54 & 61.62 & 70.70 \\ \hline
  \end{tabular}
  \vspace{-3mm} 
  
\end{table*}

The close deviation in performance metrics between the two models highlights the effectiveness of TinyLLaVa-Med's optimization, enabling it to perform at a level comparable to larger models on complex medical datasets. This supports the potential of deploying TinyLLaVa-Med in resource-constrained environments without significant loss in diagnostic accuracy.

\subsubsection{Hardware Setup and Integration}
\begin{figure}[ht]
  \centering
  \includegraphics[width=0.45\textwidth, trim={1cm 5.5cm 1cm 1cm}, clip]{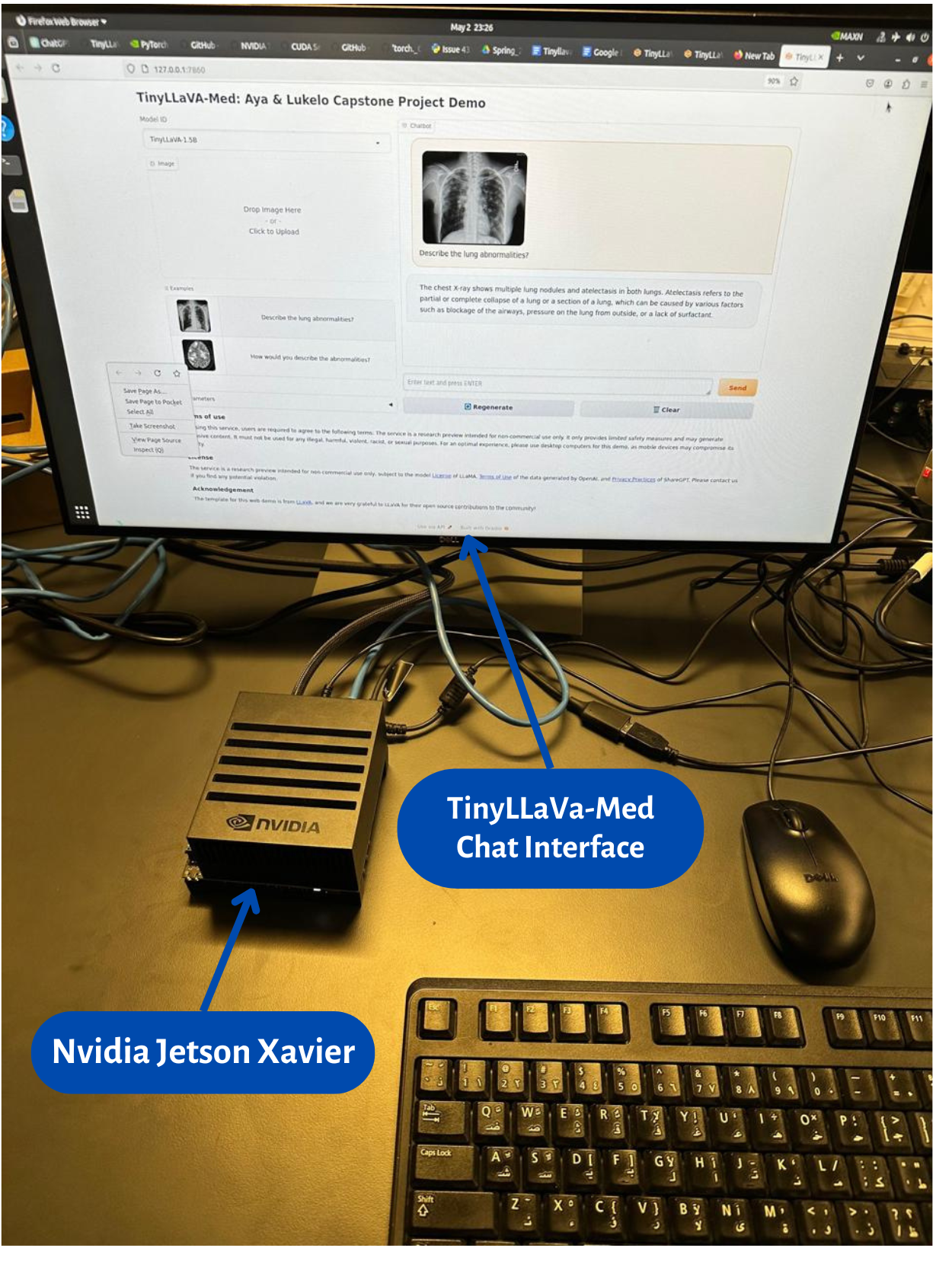}
  \caption{Hardware setup of the TinyLLaVA-Med model on NVIDIA Jetson Xavier, demonstrating the model's deployment and integration into a real-world medical environment.}
  \label{fig:hardware-setup-tinyllava-med}
\end{figure}

\begin{figure}[ht]
  \centering
  \includegraphics[width=0.5\textwidth]{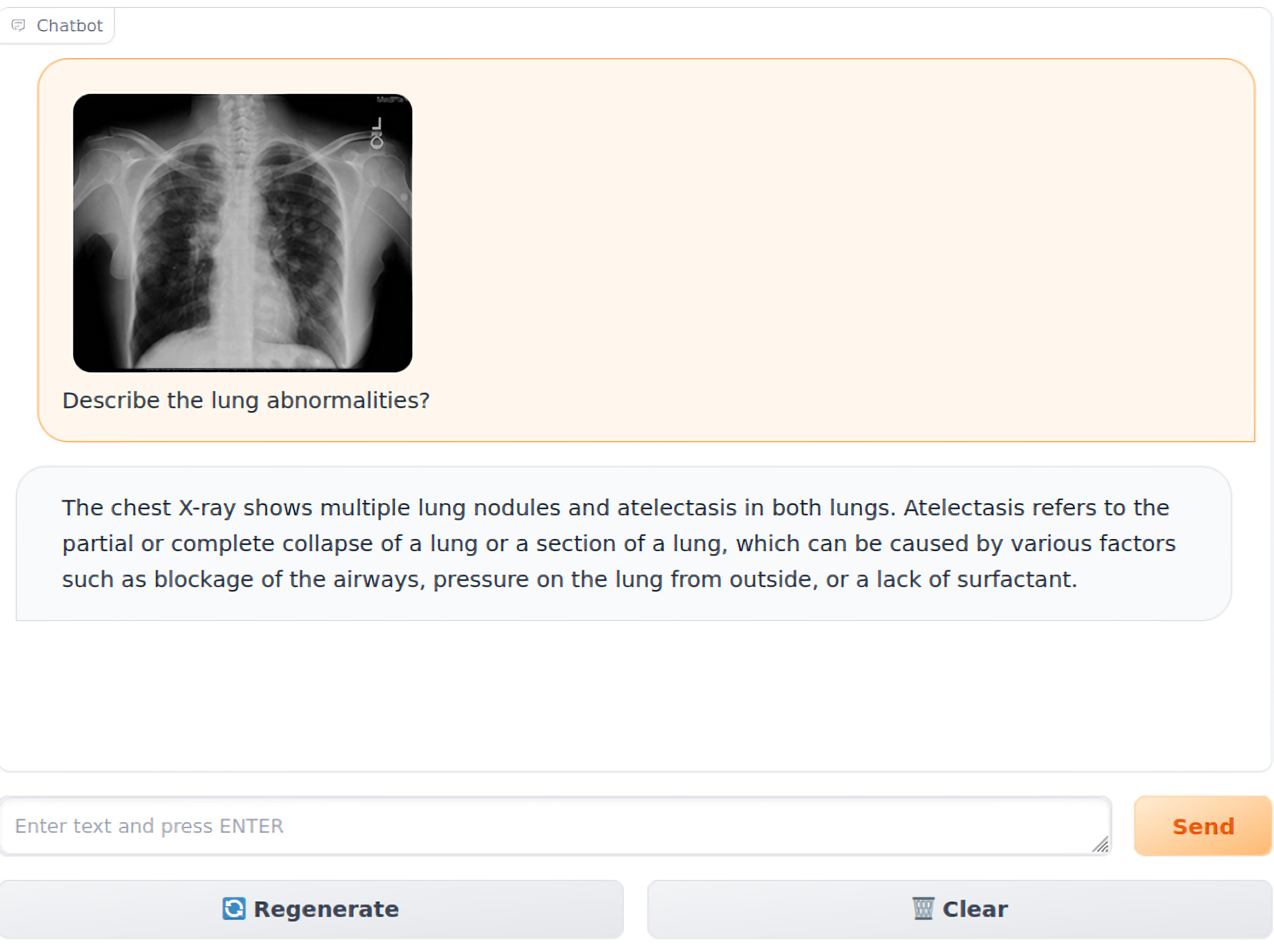}
  \caption{Close-up of the TinyLLaVA-Med chat interface deployed on NVIDIA Jetson Xavier, facilitating real-time medical image analysis.}
  \label{fig:chat-interface-tinyllava-med}
\end{figure}

Figure \ref{fig:hardware-setup-tinyllava-med} illustrates the practical deployment of the TinyLLaVA-Med model, showcasing the hardware configuration used to enable real-time medical diagnostics.This setup demonstrates how the model is integrated into a practical environment, utilizing the NVIDIA Jetson Xavier's processing power to handle the computation-intensive tasks of analyzing medical images. The workstation shown includes a monitor displaying the TinyLLaVA-Med chat interface (Figure \ref{fig:chat-interface-tinyllava-med}), where users can interact with the model, submit queries with uploaded images, and receive responses. This configuration is crucial for testing and demonstrating the model's capabilities in real-time, highlighting the practical application of our solution in medical diagnostics.

\section{DISCUSSION}

This project addresses the deployment of Multimodal Large Language Models (MLLMs) in healthcare, particularly the adaptation required for embedded systems such as Nvidia Jetson Xavier/Orin. It is crucial to understand the implications and limitations of our work and to explore pathways that further the democratization of MLLMs in the healthcare domain.

\textbf{Lack of Benchmark for MLLMs Optimization in embedded systems:}
The field of deploying MLLMs in healthcare on embedded systems is relatively unexplored, with traditional MLLMs typically targeting well-resourced environments that prioritize maximum accuracy and throughput. As a result, we faced a challenge in identifying existing benchmarks to effectively assess the optimizations of MLLMs for resource-constrained development. Consequently, we relied on the maximum capabilities of our environment, the Nvidia Jetson Xavier, as the benchmark. However, the work of other models like MoE-TinyMed \cite{jiang2024moe} suggests that this area is beginning to be more thoroughly explored. Potentially, our work could also serve as a foundational effort for establishing benchmarks in this field.

\textbf{Performance Metrics Considered:}
In developing TinyLLaVA-Med, we utilized a dual-focused set of performance metrics that cater not only to the technical demands of deployment on constrained devices but also to the requirements of medical diagnostics. These metrics include GPU utilization efficiency, energy consumption, memory footprint, and medical diagnostic accuracy. This comprehensive framework ensures that TinyLLaVA-Med is not just an effective doctor assistant MLLM but also a practical deployable tool for healthcare professionals, capable of operating effectively within the limited resources typical of remote or underserved areas. Our balanced approach ensures that the model delivers reliable medical insights with optimal power and resource efficiency, making sophisticated medical AI technology both effective and accessible.

\textbf{Implications and Contributions to the Field:}
By successfully deploying TinyLLaVA-Med, we have demonstrated that it is feasible to utilize advanced MLLMs in settings far removed from typical high-resource environments. This opens up new possibilities for medical diagnostics and treatment in locations that previously could not leverage modern AI due to infrastructural limitations. Our work contributes to the field by providing a reference framework for future projects aiming to implement high-performance AI in similar contexts. Additionally, the optimization techniques and performance metrics we have utilized can serve as a baseline for other researchers aiming to bring AI capabilities to resource-constrained environments for the healthcare domain.

\textbf{Future Research Directions:} 
To advance the deployment of MLLMs like TinyLLaVA-Med in healthcare, it is essential to address key challenges and leverage interdisciplinary collaboration for improving model performance and practical application. While TinyLLaVA-Med demonstrated robust performance in closed-ended questions, its handling of open-ended questions revealed some limitations. Open-ended questions are very important in the healthcare domain as they can provide deeper insights and comprehensive diagnostic information. However, improving the model's accuracy in this area is challenging due to the complexity of the healthcare domain. A deeper analysis of these errors and areas of improvement in this model can best be achieved by relying on the medical expertise of healthcare professionals. This analysis would also help to further refine our approach in instruction-tuning through rich and diverse dataset selection, enhancing the model's conversational capabilities and diagnostic accuracy.\\
Furthermore, the medical expertise of healthcare professionals would be vital in guiding the development of evaluation metrics that reflect both machine learning performance and healthcare standards, ensuring the models are assessed with criteria relevant to both fields. Since the expertise of healthcare professionals is crucial, more efforts are needed to promote collaboration and bridge the gap between AI technologies and the healthcare system. This collaboration will encourage more hospitals to share their data, aiming to develop models that are both more efficient and accurate. However, it is crucial to maintain data privacy and patient confidentiality to protect sensitive patient information and comply with legal and ethical standards.\\
Lastly, addressing integration issues requires robust encryption methods, adherence to healthcare data regulations, and effective strategies for data augmentation and cleaning. Given the success of this project, future research should focus on integrating other MLLM architectures to reduce computational demands while maintaining or enhancing diagnostic accuracy. Continued exploration into model compression techniques is necessary to enable sophisticated models to operate on even more limited hardware than the Nvidia Jetson Xavier/Orin. Additionally, investigating the long-term impact of deploying these models in clinical settings is crucial, including examining user acceptance, workflow integration, and overall improvements in patient outcomes. By addressing these directions, we can enhance the reliability, efficiency, and practical applicability of MLLMs in healthcare, ultimately improving patient care in resource-constrained environments. 
\vspace{-0.01em} 
\section{CONCLUSION}
Artificial Intelligence (AI) in healthcare has revolutionized the way medical professionals diagnose and treat diseases. In this field, our work with TinyLLaVA-Med addresses the critical need for advanced healthcare technologies that are accessible in low-resource settings, especially in remote and underserved areas. By successfully deploying TinyLLaVA-Med on the Nvidia Jetson Xavier, we demonstrate that it is feasible to implement sophisticated Multi-Modal Large Language Models (MLLMs) with limited computational resources without compromising diagnostic effectiveness. Our model, fine-tuned on specialized medical datasets and rigorously benchmarked on VQA-RAD and SLAKE datasets, proves that high diagnostic accuracy can be sustained even on low-performance computing platforms. This enables real-time, reliable medical decision-making capabilities in regions where advanced healthcare technology was previously inaccessible. By optimizing the performance to meet the constraints of embedded systems, TinyLLaVA-Med not only enhances healthcare delivery but also democratizes access to life-saving diagnostics.

\section*{Acknowledgement}
This work was partially supported by the NYUAD Center for Artificial Intelligence and Robotics (CAIR), funded by Tamkeen under the NYUAD Research Institute Award CG010.

\bibliographystyle{IEEEtran}
\bibliography{biblio}

\begin{thebibliography}{10}
\providecommand{\url}[1]{#1}
\csname url@samestyle\endcsname
\providecommand{\newblock}{\relax}
\providecommand{\bibinfo}[2]{#2}
\providecommand{\BIBentrySTDinterwordspacing}{\spaceskip=0pt\relax}
\providecommand{\BIBentryALTinterwordstretchfactor}{4}
\providecommand{\BIBentryALTinterwordspacing}{\spaceskip=\fontdimen2\font plus
\BIBentryALTinterwordstretchfactor\fontdimen3\font minus \fontdimen4\font\relax}
\providecommand{\BIBforeignlanguage}[2]{{%
\expandafter\ifx\csname l@#1\endcsname\relax
\typeout{** WARNING: IEEEtran.bst: No hyphenation pattern has been}%
\typeout{** loaded for the language `#1'. Using the pattern for}%
\typeout{** the default language instead.}%
\else
\language=\csname l@#1\endcsname
\fi
#2}}
\providecommand{\BIBdecl}{\relax}
\BIBdecl

\bibitem{bekbolatova2024transformative}
M.~Bekbolatova, J.~Mayer, C.~W. Ong, and M.~Toma, ``Transformative potential of ai in healthcare: Definitions, applications, and navigating the ethical landscape and public perspectives,'' in \emph{Healthcare}, vol.~12, no.~2.\hskip 1em plus 0.5em minus 0.4em\relax MDPI, 2024, p. 125.

\bibitem{poalelungi2023advancing}
D.~G. Poalelungi, C.~L. Musat, A.~Fulga, M.~Neagu, A.~I. Neagu, A.~I. Piraianu, and I.~Fulga, ``Advancing patient care: how artificial intelligence is transforming healthcare,'' \emph{Journal of personalized medicine}, vol.~13, no.~8, p. 1214, 2023.

\bibitem{jiang2017artificial}
F.~Jiang, Y.~Jiang, H.~Zhi, Y.~Dong, H.~Li, S.~Ma, Y.~Wang, Q.~Dong, H.~Shen, and Y.~Wang, ``Artificial intelligence in healthcare: past, present and future,'' \emph{Stroke and vascular neurology}, vol.~2, no.~4, 2017.

\bibitem{shaheen2021applications}
M.~Y. Shaheen, ``Applications of artificial intelligence (ai) in healthcare: A review,'' \emph{ScienceOpen Preprints}, 2021.

\bibitem{acosta2022multimodal}
J.~N. Acosta, G.~J. Falcone, P.~Rajpurkar, and E.~J. Topol, ``Multimodal biomedical ai,'' \emph{Nature Medicine}, vol.~28, no.~9, pp. 1773--1784, 2022.

\bibitem{mesko2023impact}
B.~Mesk{\'o}, ``The impact of multimodal large language models on health care’s future,'' \emph{Journal of Medical Internet Research}, vol.~25, p. e52865, 2023.

\bibitem{li2024llava}
C.~Li, C.~Wong, S.~Zhang, N.~Usuyama, H.~Liu, J.~Yang, T.~Naumann, H.~Poon, and J.~Gao, ``Llava-med: Training a large language-and-vision assistant for biomedicine in one day,'' \emph{Advances in Neural Information Processing Systems}, vol.~36, 2024.

\bibitem{singhal2023towards}
K.~Singhal, T.~Tu, J.~Gottweis, R.~Sayres, E.~Wulczyn, L.~Hou, K.~Clark, S.~Pfohl, H.~Cole-Lewis, D.~Neal \emph{et~al.}, ``Towards expert-level medical question answering with large language models,'' \emph{arXiv preprint arXiv:2305.09617}, 2023.

\bibitem{moor2023med}
M.~Moor, Q.~Huang, S.~Wu, M.~Yasunaga, Y.~Dalmia, J.~Leskovec, C.~Zakka, E.~P. Reis, and P.~Rajpurkar, ``Med-flamingo: a multimodal medical few-shot learner,'' in \emph{Machine Learning for Health (ML4H)}.\hskip 1em plus 0.5em minus 0.4em\relax PMLR, 2023, pp. 353--367.

\bibitem{eslami2023pubmedclip}
S.~Eslami, C.~Meinel, and G.~De~Melo, ``Pubmedclip: How much does clip benefit visual question answering in the medical domain?'' in \emph{Findings of the Association for Computational Linguistics: EACL 2023}, 2023, pp. 1181--1193.

\bibitem{zhang2023biomedclip}
S.~Zhang, Y.~Xu, N.~Usuyama, H.~Xu, J.~Bagga, R.~Tinn, S.~Preston, R.~Rao, M.~Wei, N.~Valluri \emph{et~al.}, ``Biomedclip: a multimodal biomedical foundation model pretrained from fifteen million scientific image-text pairs,'' \emph{arXiv preprint arXiv:2303.00915}, 2023.

\bibitem{jiang2024moe}
S.~Jiang, T.~Zheng, Y.~Zhang, Y.~Jin, and Z.~Liu, ``Moe-tinymed: Mixture of experts for tiny medical large vision-language models,'' \emph{arXiv preprint arXiv:2404.10237}, 2024.

\bibitem{zhou2024tinyllava}
B.~Zhou, Y.~Hu, X.~Weng, J.~Jia, J.~Luo, X.~Liu, J.~Wu, and L.~Huang, ``Tinyllava: A framework of small-scale large multimodal models,'' \emph{arXiv preprint arXiv:2402.14289}, 2024.

\bibitem{zhang2023large}
S.~Zhang, Y.~Xu, N.~Usuyama, J.~Bagga, R.~Tinn, S.~Preston, R.~Rao, M.~Wei, N.~Valluri, C.~Wong \emph{et~al.}, ``Large-scale domain-specific pretraining for biomedical vision-language processing,'' \emph{arXiv preprint arXiv:2303.00915}, vol.~2, no.~3, p.~6, 2023.

\bibitem{achiam2023gpt}
J.~Achiam, S.~Adler, S.~Agarwal, L.~Ahmad, I.~Akkaya, F.~L. Aleman, D.~Almeida, J.~Altenschmidt, S.~Altman, S.~Anadkat \emph{et~al.}, ``Gpt-4 technical report,'' \emph{arXiv preprint arXiv:2303.08774}, 2023.

\bibitem{lau2018dataset}
J.~J. Lau, S.~Gayen, A.~Ben~Abacha, and D.~Demner-Fushman, ``A dataset of clinically generated visual questions and answers about radiology images,'' \emph{Scientific data}, vol.~5, no.~1, pp. 1--10, 2018.

\bibitem{liu2021slake}
B.~Liu, L.-M. Zhan, L.~Xu, L.~Ma, Y.~Yang, and X.-M. Wu, ``Slake: A semantically-labeled knowledge-enhanced dataset for medical visual question answering,'' in \emph{2021 IEEE 18th International Symposium on Biomedical Imaging (ISBI)}.\hskip 1em plus 0.5em minus 0.4em\relax IEEE, 2021, pp. 1650--1654.

\end{thebibliography}

\end{document}